# Rank Correlation Measure: A Representational Transformation for Biometric Template Protection

Zhe Jin, Yen-Lung Lai
Lee Kong Chian Faculty of Engineering and Science
Universiti Tunku Abdul Rahman
Kuala Lumpur, Malaysia
jinzhe@utar.edu.my

Andrew Beng Jin Teoh
School of Electrical and Electronic Engineering
College of Engineering, Yonsei University
Seoul, South Korea
bjteoh@yonsei.ac.kr

*Abstract*— Despite a variety of theoretical-sound techniques have been proposed for biometric template protection, there is rarely practical solution that guarantees non-invertibility, cancellability, non-linkability and performance simultaneously. In this paper, a ranking-based representational transformation is proposed for fingerprint templates. The proposed method transforms a real-valued feature vector into index code such that the pairwise-order measure in the resultant codes are closely correlated with rank similarity measure. Such a ranking based technique offers two major merits: 1) *Resilient to noises/perturbations in numeric values*; and 2) *Highly nonlinear embedding based on partial order statistics*. The former takes care of the accuracy performance mitigating numeric noises/perturbations while the latter offers strong non-invertible transformation via nonlinear feature embedding from Euclidean to Rank space that leads to toughness in inversion. The experimental results demonstrate reasonable accuracy performance on benchmark FVC2002 and FVC2004 fingerprint databases, thus confirm the proposition of the rank correlation. Moreover, the security and privacy analysis justify the strong capability against the existing major privacy attacks.

## I. INTRODUCTION

Biometrics has been integrated in large-scale personal verification and identification systems. The rapid proliferation of biometric applications leads to a large number of databases that stores the biometric templates. Public worries about the security and privacy of biometric template if stolen or compromised. Such concerns are attributed to the strong binding of individuals and privacy, and further complicated by the fact that biometric traits are irrevocable. Given the above threats, a number of proposals have been reported for protecting biometric templates. It remains as a challenge task for designing a biometric template protection (BTP) scheme with the following criteria [1, 2]:

- **Non-linkability/Diversity**. It should be computationally hard to differentiate multiple instances of the protected biometric reference derived from the same biometric trait. Non-linkability prevents the cross-matching across different applications.

- **Cancelability**. A new template can be reissued once the old template is compromised.

- **Non-invertibility/Irreversibility**. It should be computationally infeasible to derive the original biometric template from the protected template and/or the helper data.

- **Performance preservation**. The accuracy performance of a protected system should be preserved or improved.

Generally, the approaches available in literature can be broadly divided into two categories: *feature transformation* (or *cancellable biometrics*) and *biometric cryptosystems* (or helper data method). Biometric cryptosystem serves the purpose of either securing the cryptographic key using biometric feature (key binding) or directly generating the cryptographic key from biometric feature (key generation) [2]. On the other hand, cancellable biometrics [3] is truly meant designed for biometric template protection. It refers to the irreversible transform applied on the biometric template to generate (and store) the transformed template such that ensuring the security and privacy of the original biometric template. If a cancellable biometric template is compromised, a new template can be re-generated from the same biometrics. The schemes of cancellable biometrics in literature vary according to different biometric modalities.

### A. Literature Review

In the past decades, numerous biometric template protection schemes have been proposed for security and privacy protection. Several decent review papers exist in this topic such as [2][4][5][6]. Another inspiring work reported by Nandakumar & Jain [7] recently, where the performance gap between theory and practice is focused while the promising direction is highlighted to bridge the gap. We refer the readers to explore more details about biometric template security from the aforementioned review papers. In this paper, we focus a few most recent and relevant fingerprint related protection schemes.

Ferrara et al. [8] proposed a recovery algorithm to reveal the original minutiae from the minutia cylinder-code (MCC) template, a state-of-the-art fingerprint descriptor proposed by Cappelli et al. [9]. A non-invertible scheme on MCC is hence proposed, namely protected minutia cylinder-code (P-MCC) by using binary principle component analysis. Although, the cancellability is not addressed in original P-MCC, later, a two-factor protection scheme on P-MCC, namely 2P-MCC [10] is proposed to make the protected MCC template cancellable. The specific design of MCC and its variations for point set data limits the usage in other popular biometric modalities like face and iris.

Recently, a more generalized template protection technique, namely bloom filter has been introduced for iris [11], face [12] and fingerprint [13] respectively. Bloom filter demonstrates a well adaption for the popular biometric modalities. However, despite the decent performance preservation, the security and privacy of bloom filter based schemes remains unresolved. For instance, Hermans et al. [14] demonstrated a simple and effective attack scheme that matches two protected templates derived from a same IrisCode using different secret bit vectors, thus break the requirement of non-linkability. Moreover, a security analysis also reported reveals that the false positive or recover the key can be accomplished with a low attack complexity of $2^{25}$ and $2^2$ to $2^8$ attempts respectively [14]. Bringer et al. [15] further analyzed the non-linkability of the protected templates generated from two different IrisCode of the same subject. They revealed when the key space is too small, a brute force attack would be succeeded while the identification/authentication rate declines if the key size is increased. Experiment confirms the vulnerability of irreversibility with block width of 16 or 32 in Bloom filter scheme.

Sandhya & Prasad [16] proposed a k-nearest neighbor structure from fingerprint minutia to construct a fixed-length binary representation that is invariant to rotation and translation. A discrete Fourier transform is applied on the binary representation to result a complex vector that is multiplied with Gaussian random matrix to obtain a cancellable template. This technique can achieve reasonable recognition accuracy. However, the security of the proposed method is insufficiently analyzed. Meanwhile, Wang and Hu [17] proposed a blind system identification approach to protect biometric template. This is motivated by fact that source signal cannot be recovered if the identifiability is dissatisfied in blind system identification. This new approach exhibits well accuracy performance preservation and the irreversibility of transformed template is justified theoretically and experimentally. However, the protected template against other major attacks (e.g. ARM) is unknown.

In this paper, we report a generic method for biometric template protection based on rank correlation measure. The proposed method is inspired from the "Winner Takes All" hashing [18] that originally meant for solving the fast similarity search problem. In connection with biometric template protection, several alternations have to be made to meet the specific requirements of biometric template protection. In short, the protected template is able to (1) conceal the information of the original biometric features reliably; (2) satisfy the performance preservation and cancellability; and (3) incredibly easy for practical applications. We demonstrate the feasibility of this method with fingerprint modality.

*B. Preliminary*

The current section describes the "Winner Takes All" (WTA) on which our work is based. Yagnik et al. [18] introduced WTA hashing for fast similarity search. The basic idea of WTA is to compute the ordinal embedding of an input feature based on the partial order statistics. More specifically, WTA is a non-linear transformation based on the implicit order rather than the absolute/numeric values of the input feature, and therefore, offers certain degree of resilience to numeric noise while giving a good indication of inherent similarity between the compared items/vectors [18]. The overall WTA hashing procedure can be summarized into four steps as shown in Fig. 1:

- Perform $H$ random permutations on the input feature $\mathbf{X} \in \mathbb{R}^n$, $n$ is the dimension of the input feature $\mathbf{X}$.
- Select the first $K$ (i.e. window size) items of the permuted input feature.
- Choose the largest element within the $K$ items.
- Record the corresponding index value using $\log_2 K$ bits.

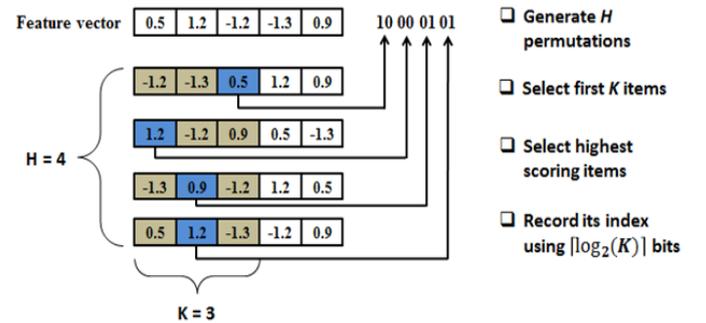

Fig. 1: Overall procedure of WTA (adopted from [18]).

## II. PROPOSED METHOD

Going along with WTA, in this Section, we present a rank-based representational transformation for biometric template protection. The overall transformation is illustrated in Fig 2. Assume an input feature vector $\mathbf{X} \in \mathbb{R}^n$ and $m$ independent hash functions $H(\mathbf{X}) = \{h_i(\mathbf{X}) | i = 1, \dots, m\}$ where each hash function consists of $p$ degree of polynomial kernel. $m$ and $p$ are determined empirically. The hashed code is generated by concatenating the outputs from $m$ independent hash functions, therefore, the length of protected template is $m$. The procedure of the proposed method is described as follows:

- **Random Permutation**: For each $H(\mathbf{X})$, generating a permutation set $\theta$ consists of $p$ randomly generated permutation seeds. Then the input feature vector $\mathbf{X}$ is permuted using the permutation seeds to generate permuted template, $\widehat{\mathbf{X}} = \{\widehat{\mathbf{X}}_j \in \mathbb{R}^n | j = 1, \dots, p\}$.

- **Hardamard Product-code Generation**: Generate the $p^{th}$ ordered Hardamard product code by taking the element-wise multiplication of all the permuted templates that can be described as $\overline{\mathbf{X}} = \prod_{j=1}^{p}(\widehat{\mathbf{X}}_j)$. In this manner, $p$ is regarded as a $p^{th}$ degree of polynomial kernel.

- For each product code, $\overline{\mathbf{X}}$ select the first $k$ elements, while $1 < k < n$.

- Record the index of the maximum value in the first $k$ elements (or $k$-size window), and denote it as $t$.

- Repeat step 1 to 4 using different permutations set $\theta_{(l,i)}$, where $l \in [1,p]$, $i \in [1,m]$ to form a $1 \times m$ hashed code $T = \{t_i \in \mathbb{Z}^n | i = 1, ..., m\}$, $T \in [0, k]$.

In the event of template compromise, $m$ independent hash functions, $H(\mathbf{X}) = \{h_i(\mathbf{X}) | i = 1, ..., m\}$ are regenerated with different random permutation seeds to replace the compromised hash functions. A new template can be re-issued by repeating the steps 1 to 5 described above. The effectiveness of cancellability is experimentally verified in Section 4.3. In real world scenario, the permutation seed is user-specific for cancellability. However, lost token/seed scenario should be focused as it is closely associated to accuracy performance, security and privacy attacks [1, 19]. To evaluate the lost token scenario, our experiment is performed with same permutation seed for all subjects. The accuracy performance in lost token scenario is presented in Section 4.2.

| Algorithm 1. Rank Hashing with Polynomial Kernel |
|---|
| **Input:** Feature vector **X**, $p$ degree of polynomial kernel, number of random permutation $m$, window size $k$ |
| *For each polynomial kernel $\theta_{(l,i)}$, $i \in [1,m]$, $l \in [1,p]$* |
| **Step 1**: Permute elements in **X** based on $\theta_{(i,l)}$, $\widehat{\mathbf{X}} = perm(\mathbf{X})$, $perm(.)$ refers random permutation function |
| **Step 2**: Initialize $i^{th}$ hashed code $t_i = 0$ |
| **Step 3**: Hardamard product-code generation and output hashed codes<br>For $j=1: k$<br>    Set $\overline{\mathbf{X}}(j) = \prod_{l=1}^{p}(\widehat{\mathbf{X}}_l(j))$<br>    if $\overline{\mathbf{X}}(j) > \overline{\mathbf{X}}(t_i)$ then $t_i = j$<br>End for |
| **Output:** Hashed code $T = \{t_i | i = 1, ..., m\}$ and $T \in [1, k]$ |

*A. Matching*

Briefly, rank correlation refers to the measurement of ordinal association. In this context, rank correlation measure (also known as ordinal measure) is based on the relative ordering of values in a given range. It is known as a useful measurement for pixel correspondence in stereo matching [20]. Furthermore, rank correlation has been well studied as similarity measurement between two feature representations, which is defined by the degree to which their feature dimension rankings agree [18]. The pairwise-order, $PO$ is the simplest similarity measure for rank correlation [18]. The exposition of rank correlation and PO are given in [21].

Assume two hashed codes, enrolled $\boldsymbol{T}^e = \{t_i^e | i = 1, ..., m\}$ and query $\boldsymbol{T}^q = \{t_j^q | j = 1, ..., m\}$). The *PO* function can be defined as follows:

$$PO(\boldsymbol{T}^e, \boldsymbol{T}^q) = \sum_i \sum_{j<i} Q\left((\boldsymbol{T}_i^e - \boldsymbol{T}_j^e)(\boldsymbol{T}_i^q - \boldsymbol{T}_j^q)\right)$$
$$= \sum_i R_i(\boldsymbol{T}^e, \boldsymbol{T}^q)$$

where $R_i(\boldsymbol{T}^e, \boldsymbol{T}^q) = |L(\boldsymbol{T}^e, i) \cap L(\boldsymbol{T}^q, i)|$
$L(\boldsymbol{T}^e, i) = \{j | \boldsymbol{T}^e(i) > \boldsymbol{T}^e(j)\}$
$L(\boldsymbol{T}^q, i) = \{j | \boldsymbol{T}^q(i) > \boldsymbol{T}^q(j)\}$
$Q(z) = \begin{cases} 1 & z > 0 \\ 0 & z \leq 0 \end{cases}$

The operational PO function can be reformulated as [18]:

$$S_k(\boldsymbol{T}^e, \boldsymbol{T}^q) = \frac{\sum_{i=0}^{n-1}\binom{R_i(\boldsymbol{T}^e, \boldsymbol{T}^q)}{k-1}}{\binom{n}{k}}$$

According to the rank correlation measure above, the proposed method essentially follows the intrinsic nature of PO measure (e.g. highest order selection). Thus, this similarity measure can be used to compute the similarity score of two distinct biometric features after ranking-based transformation as described in Algorithm 1. Mathematically, $S_k(\boldsymbol{T}^e, \boldsymbol{T}^q)$ represents the *probability of collision* given by a random permutation $\theta$ on hashed codes $\boldsymbol{T}^e, \boldsymbol{T}^q$. In this context, a collision, $c$ refers to the maximum value appeared in the identical position for both $\boldsymbol{T}^e, \boldsymbol{T}^q$ within a $k$-size window, i.e. $c \leftarrow \{\boldsymbol{T}_i^e = \boldsymbol{T}_i^q\}, i = 1, ..., k$. The higher probability of collision implies higher similarity of $\boldsymbol{T}^e, \boldsymbol{T}^q$ and vice-versa. In our implementation, the number of collision is calculated by counting the number of 0 after the element-wise subtraction of $\boldsymbol{T}^e, \boldsymbol{T}^q$ is performed.

### III. EXPERIMENTS AND DISCUSSIONS

In this paper, a real-valued and fixed-length fingerprint vector with length 299 that generated from MCC and KPCA [22] is used as input data to evaluate the proposed method. We refer the readers for the details about the fingerprint vector construction in [23]. The evaluations are conducted on six public fingerprint datasets, FVC2002 (DB1, DB2, DB3) [24] and FVC2004 (DB1, DB2, DB3) [24]. Each dataset consists of 100 users with 8 samples per user. In total, there are 800 (100×8) fingerprint images in each dataset. The performance accuracy of the proposed method is assessed using Equal Error Rate (EER) and the genuine-imposter distribution. Noted that since the random permutation is applied, to avoid the bias of single random permutation, the EERs is calculated by taking an average of EER repeated for 5 times.

For matching protocol, as described in [23], 1st to 3rd samples of each identity are used as training samples to generate the fixed-length fingerprint representation; thus, the rest samples (i.e. 4th – 8th) of each identity are used in this experiment. There are 500 (100×5) in total. Within this subset of data, The Fingerprint Verification Competition (FVC) [24] protocol is applied across the six data sets, which yields 1000 genuine matching scores and 4950 imposter matching scores for each data set.

## A. Effect of window size k, polynomial kernel p, and number of hashing functions m

We first investigated the effect of window size $k$ with respect to the performance in terms of EER. In this experiment, $k$ is varied from 50, 80, 100, 128, 156, 200 to 250 by leveling off $m$ ($m$=600). The identical setting is repeated for $p$=[2,3,4,5]. Then, Fig. 3 shows the curves of "EER (%)-vs-$k$" using the aforementioned parameters setting on FVC2002 DB1 and DB2. We can observe that:

- The EER drops gradually when larger $k$ is applied and levels off when $k$ becomes large. This is not surprised that smaller $k$ leads to less feature components taking into account, which leads to insufficient discriminability; while larger $k$ indicates more salient features for matching;

- The smaller $p$, the lower EER, i.e. better performance. Algorithm 1 (step 3) tells us that the Hardamard product-code is obtained by element-wise multiplication of the $p$ permuted templates, $\overline{\mathbf{X}}(j) = \prod_{l=1}^{p}(\widehat{\mathbf{X}}_l(j))$. Such an operation heightens the difficulty against inversion at the expense of introducing distortion in the product code as noises may involve in one or multiple permuted templates. Thus, it is expected that the performance drops with large $p$. This also demonstrates the common trade-off suffered in cancellable biometrics, namely performance-security trade-off.

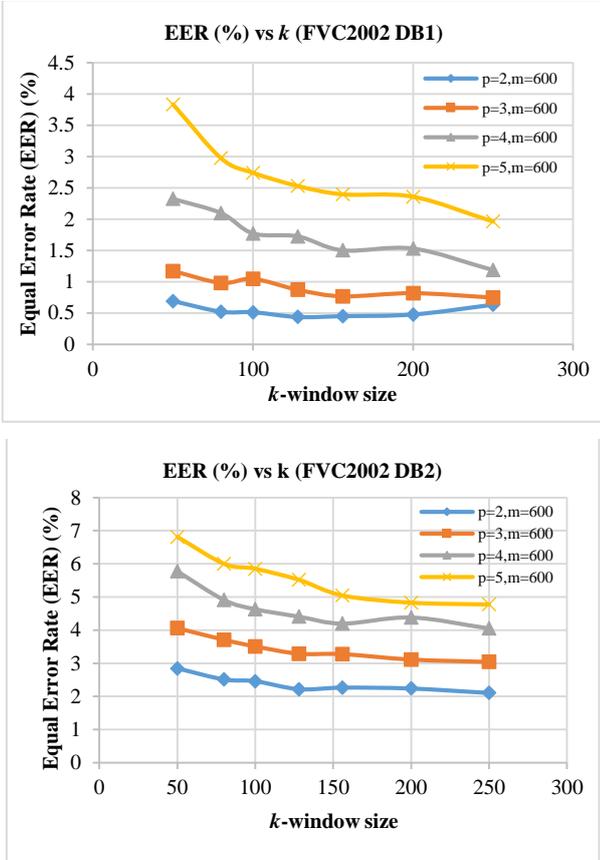

Fig. 3: The curves of "EER (%) vs $k$" on FVC2002 DB1 and DB2.

Apart from the above, we also examined the relation between the number of hashing functions $m$ and EER. Evaluation has been carried out by increasing the $m$ from 5, 10, 50, 100, 300, 500, 800 and 1000 while fixing $k$=250, and $p$=2. As expected, a better EER can be attained with respect to the increment of $m$ and level off at large $m$ as illustrated in Fig. 4. Since randomization (i.e. random permutation) is involved in the hashing functions, Central limit theorem [25] tells us that the random vector constructed is approximately Gaussian distributed with sufficiently large $m$. Experimental evidence suggests that with $m$=200 and $k$=250, we obtain good performance accuracy. This resembles a well-known instance, namely Kernelized Locality-Sensitive Hashing (KLSH) [26].

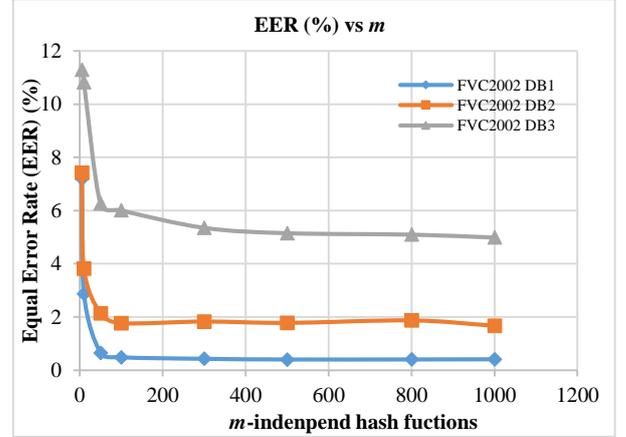

Fig. 4: The curves of "EER (%) vs $m$" on FVC2002 (DB1, DB2 and DB3).

TABLE I.  PERFORMANCE ACCURACY (LOST TOKEN/SEED SCENARIO) AND COMPARISON

| Methods | FVC2002 DB1 | FVC2002 DB2 | FVC2002 DB3 | FVC2004 DB1 | FVC2004 DB2 | FVC2004 DB3 |
|---|---|---|---|---|---|---|
| **Without Template protection** | | | | | | |
| MCC [9] | 0.60% | 0.59% | 3.91% | 3.97% | 5.22% | 3.82% |
| Fixed-length Representation [23] | 0.20% | 0.19% | 2.30% | 4.70% | 3.13% | 2.80% |
| **With Template Protection** | | | | | | |
| **Proposed** | **0.43%** | **2.11%** | **6.47%** | **4.45%** | **7.87%** | **8.29%** |
| 2P-MCC$_{64,64}$ [10] | 3.3% | 1.8% | 7.8% | 6.3% | - | - |
| Bloom Filter [27] | 2.3% | 1.8% | 6.6% | 13.4% | 8.1% | 9.7% |
| Wang & Hu [17] | 4% | 3% | 8.5% | - | - | - |

## B. Accuracy Performance Evaluation

In this section, the accuracy performance experiments on FVC2002 and FVC2004 using the best parameters found in the previous section is carried out. Table 1 presents the accuracy results as well as comparisons with the baseline system and other methods. It can be observed that:

1) the performance accuracy of the proposed method does not degrade significantly with respect to its fingerprint vector

counterpart [18]; this confirms the rank correlation after transformation;

2) the proposed method outperforms the existing fingerprint minutiae based template protection methods [10, 17, 27]; this is attributed to the superior MCC descriptor and the performance preservation traits of the ranking based transformation.

## IV. SECURITY ANALYSIS

### A. Non-invertibility Analysis

Non-invertibility in this context refers to the computational hardness in restoring the fingerprint vector from the hashed code with and without helper data. Here, we assume the adversary manages retrieve the hashed codes and he knows well hashing algorithm as well as the corresponding parameters (e.g. $m$, $k$, $p$ and permutation seeds). We noted that the proposed ranking-based transformation converts the real-valued feature into index value (i.e. hashed code). There is no clue for an adversary to guess the fingerprint vector information directly from the stolen hashed code alone without other information. The way for the adversary to attack is only to guess the real-value directly. In the worst case, assume the adversary learns the minimum and maximum values of the feature components. Let's take FVC2002 DB1 as an example, the minimum and maximum values of the feature components are -0.2504 and 0.2132 respectively. The adversary has to examine from -0.2504, -0.2503, -0.2502 and so on, until the maximum 0.2132. Thus, there are totally 4636 possibilities. In our implementation, the precision is fixed at four decimal digits, the possibility of guessing a single feature component of fingerprint vector requires 4636 attempts ($\approx 2^{12}$). Thus, the entire 299 feature components require around $2^{12 \times 299} = 2^{3588}$ attempts in total. The possibilities to correctly guess *a single* and *entire* feature components are presented in Table 2. Obviously, such possibilities are computational infeasible.

### B. Attacks via Record Multiplicity (ARM)

ARM refers to a *privacy* attack, which utilizes multiple compromised protected templates with and without knowledge and parameters that associated to the algorithm to reconstruct the original biometric template [28]. In our case, ARM is computationally hard to gain the numerical value as the stored templates have been transformed into rank space that is not correlated with original space. The complexity to gain the real-valued feature vector is presented in Table 2. The concern for our specific case is whether ARM is feasible to gain illegitimate access. Theoretically, if knowing the order of the feature components (not necessary the numerical value) and the permutation seeds, an elaborate faked feature vector can be formed. Consequently, the maximum value resulted from the product of two permuted feature vectors appears in the desired rank/position. The system thus can be broken.

Now, the complexity for ARM is shifted to the complexity of determining the order of the feature components in the fingerprint vector. For instance, let $\mathbf{X} = \{x_a, x_b, x_c\}$ be the input feature, with $p = 2$ and two randomly permuted feature vectors $\widehat{\mathbf{X}}_1 = \{x_c, x_a, x_b\}$ and $\widehat{\mathbf{X}}_2 = \{x_b, x_c, x_a\}$. The Hardamard product code $\overline{\mathbf{X}} = \{x_b x_c, x_a x_c, x_a x_b\}$ can be computed. Assume $x_a x_c$ is the largest component after Hardamard multiplication, i.e. $x_a x_c = \max\left(\prod_{l=1}^{2}\left(\widehat{\mathbf{X}}_l(j)\right)\right), j = 1,2,3$. Thus, two inequalities can be derived as follow: $x_a x_c > x_b x_c$ and $x_a x_c > x_a x_b$. We can further reason that for $x_a > x_b$ and $x_c > x_b$, the adversary can retrieve the entire order information (e.g. $x_a > x_c > x_b$) by repeating this process using multiple hashed codes. However, in practice, our feature components contain both positive and negative values. For instance, let $\mathbf{X} = \{x_a, x_b, x_c\} = \{-0.2, 0.5, -0.1\}$, by using $x_a x_c > x_b x_c$, ie. $(-0.2) \times (-0.1) > (0.5) \times (-0.1)$, we obtain $(-0.2) < (0.5)$, hence $x_a < x_b$. This is contradicted with $x_a > x_b$. Therefore, the inequality reasoning only valid under the assumption where the feature values are all with coefficient +1. We now can conclude that the inequality relation is insufficient for launching ARM effectively.

TABLE II. COMPLEXITY TO INVERT SINGLE AND ENTIRE FEATURE COMPONENTS.

| Databases | Min value with four decimal precision | Max value with four decimal precision | Possibilities for single feature component | Total possibilities for entire feature |
|---|---|---|---|---|
| FVC2002 DB1 | -0.2504 | 0.2132 | $4636 \approx 2^{12}$ | $2^{12 \times 299} = 2^{3588}$ |
| FVC2002 DB2 | -0.2409 | 0.2484 | $4893 \approx 2^{12}$ | $2^{12 \times 299} = 2^{3588}$ |
| FVC2002 DB3 | -0.1919 | 0.2372 | $4291 \approx 2^{12}$ | $2^{12 \times 299} = 2^{3588}$ |
| FVC2004 DB1 | -0.2487 | 0.1748 | $4235 \approx 2^{12}$ | $2^{12 \times 299} = 2^{3588}$ |
| FVC2004 DB2 | -0.2357 | 0.1950 | $4307 \approx 2^{12}$ | $2^{12 \times 299} = 2^{3588}$ |
| FVC2004 DB3 | -0.1947 | 0.1796 | $3742 \approx 2^{11}$ | $2^{12 \times 299} = 2^{3289}$ |

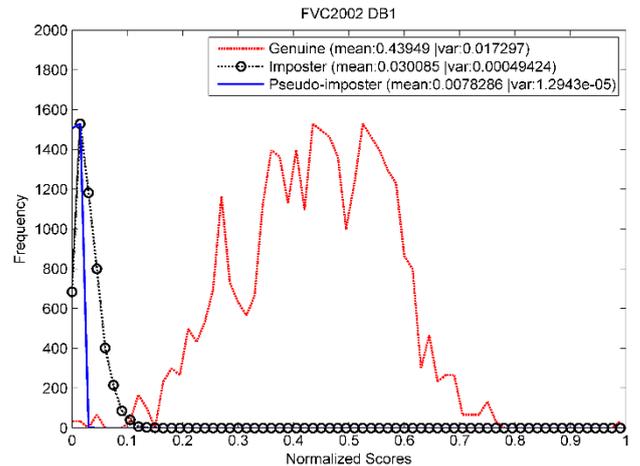

Fig. 5: The genuine, imposter, and pseudo-imposter distribution with $p = 2, k = 128, m = 600$ on FVC2002 DB1.

### C. Cancellability

In this section, cancellability is evaluated by conducting the experiment where 101 hashed codes for each fingerprint vector are generated with 101 sets distinct random permutation seeds, and then the first hashed code is matched with the other 100 hashed codes. The entire process is repeated and produces 100 × (5×100) = 50000 pseudo-imposter scores. The genuine, imposter, and pseudo-imposter distribution are computed with $p = 2, k = 128, m = 600$ are given in Fig. 5. Note that the numbers of scores are different for the imposter and pseudo-imposter matching, this is because in pseudo- imposter matching, we only focus on the matching scores between the first hashed code and the newly generated hashed code for each fingerprint vector (same user) for cancellability evaluation. From Fig. 5, a large degree of overlapping occurs between the imposter and pseudo-imposter distributions. This implies the newly generated hashed codes with the given 100 random permutation sets are distinctive even though it is generated from the identical fingerprint vector. In terms of verification performance, we obtain EER = 0.16% in which intersection of genuine and pseudo-imposter distribution is taken. This verifies that the proposed method satisfies the cancellability property requirement.

## V. Conclusion

In this paper, we presented a ranking-based transformation for fingerprint template protection. Yet, the proposed method is not only limited in fingerprint modality, other popular biometric modalities (e.g. face) with real-valued representation are also applicable. It is worthy to note that binary representation can be easily adopted by taking the position of first '1' in this method. In view of the template protection design criteria, the proposed method enables performance preservation thank to the nice property of rank correlation and strong capability against the existing major security and privacy attacks, thus comply with non-invertibility and non-linkability criteria. Experiment results suggest that cancellable property is also achieved. Moreover, the simplistic nature makes the development and deployment of the proposed method truly practical likely.